\documentclass[letterpaper, 10 pt, conference]{ieeeconf}  

\IEEEoverridecommandlockouts
\usepackage{amsmath} 
\usepackage[hidelinks]{hyperref}
\usepackage{bbding}
\usepackage{bm}
\usepackage{multirow}
\usepackage{xcolor}
\usepackage{textcomp}
\usepackage{gensymb}
\usepackage{graphicx}
\usepackage[detect-none]{siunitx}
\DeclareSIUnit{\nothing}{\relax}
\usepackage{color}
\usepackage{tabularray}
\usepackage{array}
\usepackage{booktabs}

\usepackage{amssymb}
\usepackage{pifont}
\newcommand{\cmark}{\ding{51}}%
\newcommand{\xmark}{\ding{55}}%

\definecolor{somegray}{rgb}{0.5, 0.5, 0.5}
\newcommand{\darkgrayed}[1]{\textcolor{somegray}{#1}}
\makeatletter
\newcommand*\titleheader[1]{\gdef\@titleheader{#1}}
\AtBeginDocument{%
  \let\st@red@title\@title
  \def\@title{%
    \vskip-1.5em
    \bgroup\normalfont\large\centering\@titleheader\par\egroup
    \vskip1.5em\st@red@title}
}

\makeatother

\titleheader{\darkgrayed{This paper has been accepted for publication in the IEEE ICRA 2025 \copyright 2025 IEEE.}}

\title{\LARGE \bf
A Map-free Deep Learning-based Framework for Gate-to-Gate Monocular Visual Navigation aboard Miniaturized Aerial Vehicles}

\author{Lorenzo Scarciglia$^{1}$, Antonio Paolillo$^{1}$, and Daniele Palossi$^{1,2}$
\thanks{This work has been partially funded by the Hasler Foundation (\# 23048).}%
\thanks{$^{1}$L. Scarciglia, A. Paolillo, and D. Palossi are with the Dalle Molle Institute for Artificial Intelligence (IDSIA), USI-SUPSI, Lugano, Switzerland.}%
\thanks{$^{2}$D. Palossi is also with the Integrated Systems Laboratory (IIS), ETHZ, Z\"urich, Switzerland. Contact author: lorenzo.scarciglia@supsi.ch}%
}

\begin{document}

\maketitle
\thispagestyle{empty}
\pagestyle{empty}


\begin{abstract}
Palm-sized autonomous nano-drones, i.e., sub-\SI{50}{\gram} in weight, recently entered the drone racing scenario, where they are tasked to avoid obstacles and navigate as fast as possible through gates.
However, in contrast with their bigger counterparts, i.e., \SI{}{\kilo\gram}-scale drones, nano-drones expose three orders of magnitude less onboard memory and compute power, demanding more efficient and lightweight vision-based pipelines to win the race.
This work presents a map-free vision-based (using only a monocular camera) autonomous nano-drone that combines a real-time deep learning gate detection front-end with a classic yet elegant and effective visual servoing control back-end, only relying on onboard resources.
Starting from two state-of-the-art tiny deep learning models, we adapt them for our specific task, and after a mixed simulator-real-world training, we integrate and deploy them aboard our nano-drone.
Our best-performing pipeline costs of only \SI{24}{\mega\nothing} multiply-accumulate operations per frame, resulting in a closed-loop control performance of \SI{30}{\hertz}, while achieving a gate detection root mean square error of 1.4 pixels, on our $\sim$\SI{20}{\kilo\nothing} real-world image dataset.
In-field experiments highlight the capability of our nano-drone to successfully navigate through 15 gates in \SI{4}{\minute}, never crashing and covering a total travel distance of $\sim$\SI{100}{\meter}, with a peak flight speed of \SI{1.9}{\meter/\second}.
Finally, to stress the generalization capability of our system, we also test it in a never-seen-before environment, where it navigates through gates for more than \SI{4}{\minute}.
\end{abstract}


\section*{Supplementary material}
\noindent
Demo video available at: \href{https://youtu.be/aY0dIhoKQXg}{https://youtu.be/aY0dIhoKQXg}.


\section{Introduction} \label{sec:intro}

\begin{figure}[t]%
    \centering%
    \includegraphics[width=\columnwidth]{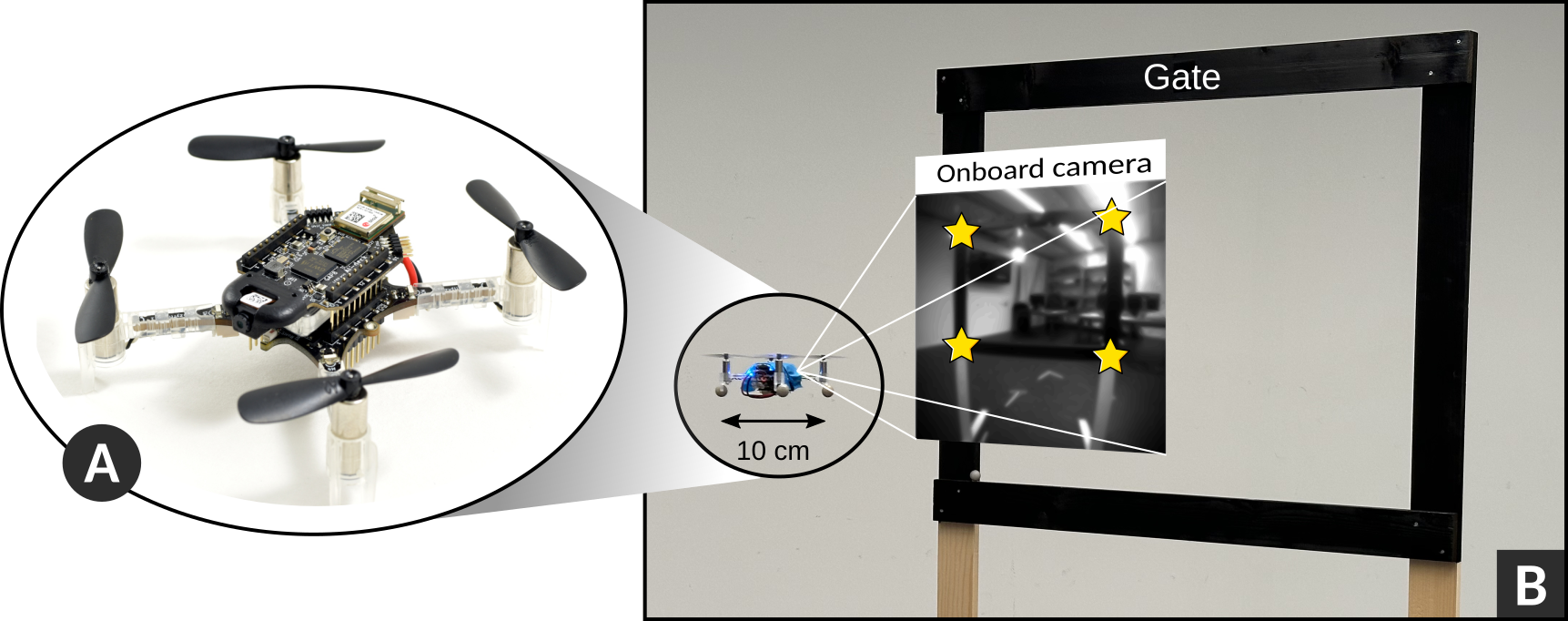}%
    \caption{A) Our autonomous nano-drone. B) Vision-based gate-to-gate navigation using the prediction of the four corners (stars).}
    \label{fig:demo}%
\end{figure}

Miniaturized autonomous robots (Fig.~\ref{fig:demo}A) are gaining ever more momentum due to their vast applicability thanks to their reduced size, i.e., $\sim$\SI{10}{\centi\meter} of diameter, and weight, i.e., less than \SI{50}{\gram}.
Miniaturized aerial drones, also called \textit{nano-drones}, can inspect and monitor narrow and inaccessible places for bigger drones~\cite{duisterhof2021sniffy, lamberti2024distilling}; they can safely operate nearby humans~\cite{gomes2016bitdrones, palossi2021fully}, embodying the human-robot interaction paradigm~\cite{sheridan2016human}, and are extremely cheap, compared to \SI{}{\kilo\gram}-scale counterparts, due to their simplified electronic~\cite{giernacki2017crazyflie}.

In the last few years, nano-drones also entered the autonomous drone racing scenario~\cite{lamberti2024sim}, where robots are tasked to autonomously, i.e., without relying on any external infrastructure or off-board computational aids, navigate in a challenging environment passing through gates and dodging static and dynamic obstacles.
As in any racing domain, drone competitions act as a proxy to foster and push the technology beyond the current state-of-the-art (SoA), requiring always more agile and intelligent cyber-physical systems.
Additionally, nano-drone races offer the opportunity to face novel challenges, mainly due to the strong limitations of these platforms, such as \textit{i}) the onboard computing power (1000$\times$ less than in standard-sized racing drones, e.g., equipped with NVIDIA Jetson/Xavier boards), \textit{ii}) the rudimentary sensors available (e.g., low-resolution monochrome cameras), and \textit{iii}) their ultra-tight memory budget (up to a few \SI{}{\mega\byte} compared to tens of \SI{}{\giga\byte} available in top-notch racing drones).

As a consequence of all these limitations, even the winning autonomous nano-drone~\cite{lamberti2024sim}, competing in the last ``Nanocopter AI Challenge'' hosted during the International Micro Air Vehicles (IMAV) conference, limited its onboard task only to reactive obstacle avoidance.
The secondary task of gate-based navigation was ignored due to the complexity of designing and deploying aboard the nano-drone an effective yet lightweight solution.
In this light, \textbf{our work marks a clear step forward by designing, deploying, and field-testing a novel deep learning-based (DL) monocular visual servoing system capable of running in real-time within the limited resources of a nano-drone and precisely performing the gate-based navigation task.}
Our solution is based on a DL front-end for gate detection predicting the gate corner coordinates in the image space, as depicted in Fig.~\ref{fig:demo}B.
Then, these predictions are fed back to a classic image-based visual servoing (IBVS) back-end~\cite{chaumette2006vs} producing control signals, i.e., the drone's 3D linear velocity and the vertical angular one, for the low-level controller of the drone, i.e., a proportional–integral–derivative (PID) control cascade.

\begin{table*}[t]
\centering
\caption{Micro/nano-sized drones performing obstacle avoidance (OA) and/or gate-based navigation (GN) employing deep learning-based (DL) or geometry-based algorithms with both sensing and computing devices and their power consumption.}
\label{tab:soa}
\begin{tblr}{
 width = \linewidth,
 colspec = {Q[175] Q[50] Q[70] Q[230] Q[90] Q[90] Q[180] Q[130]},
 colspec={lccccccc},
 hline{1,2,13} = {-}{0.08em},
}
\textbf{Reference} & \textbf{Class} & \textbf{Task} & \textbf{Sensors} & \textbf{Algorithm} & \textbf{Map-free} & \textbf{Computing device} & \textbf{Power (sensors+computing)} \\
Jung et al.~\cite{jung2018direct} & Micro & GN/OA & Stereo camera + LiDAR & Geometry & \xmark & NVIDIA TK1 & 3.2+\SI{5}{\watt} \\
Jung et al.~\cite{jung2018perception} & Micro & GN & Stereo camera + LiDAR & DL & \cmark & NVIDIA Jetson TX2 & 3.2+\SI{15}{\watt} \\
Kaufmann et al.~\cite{kaufmann2019beauty} & Micro & GN & Monocular camera  & DL & \xmark & Intel UpBoard & N.D.+\SI{13}{\watt} \\
De Wagter et al.~\cite{de2021artificial} & Micro & GN & Stereo camera + LiDAR & DL  & \xmark & NVIDIA Xavier & 3.8+\SI{30}{\watt} \\
Foehn et al.~\cite{foehn2022alphapilot} & Micro & GN & Stereo camera + LiDAR & DL & \xmark & NVIDIA Xavier & 3.8+\SI{30}{\watt} \\
Pham et al.~\cite{pham2022pencilnet} & Micro & GN & Camera + Tracking camera & DL & \xmark & NVIDIA Jetson TX2 & 4.5+\SI{15}{\watt} \\
Yang et al.~\cite{yang2024lightweight} & Micro & GN/OA & Depth camera & Geometry & \xmark & NVIDIA Jetson NX & 3.2+\SI{20}{\watt} \\
Li et al.~\cite{li2020visual} & Nano & GN & Monocular camera (1.3MP) & Geometry & \xmark & JeVois A33 & N.D.+\SI{3.5}{\watt} \\
Kalenberg et al.~\cite{kalenberg2024stargate} & Nano & GN/OA & ULP camera + ToF & DL & \cmark & GWT GAP8 & 322+\SI{100}{\milli\watt} \\
Lamberti et al.~\cite{lamberti2024sim} & Nano & OA & ULP camera & DL & \cmark & GWT GAP8 & 2+\SI{100}{\milli\watt} \\
\textbf{Ours} & \textbf{Nano} &\textbf{GN} & \textbf{ULP camera} & \textbf{DL} & \cmark & \textbf{GWT GAP8} & \textbf{2+\SI{100}{\milli\watt}} \\ 
\end{tblr}
\end{table*}

First, we select two different SoA tiny DL models for our front-end: the PULP-Frontnet convolutional neural network (CNN) originally developed for a human pose estimation task~\cite{palossi2021fully} and a fully convolutional neural network (FCNN) designed for predicting the pose of a target drone~\cite{crupi2024high}.
Then, we adapt and train both networks for our gate detection task by employing the Webots simulator at first ($\sim$\SI{600}{\kilo\nothing} images) and fine-tuned them using only $\sim$\SI{70}{\kilo\nothing} real-world images collected in our laboratory.
On our custom real-world testing dataset ($\sim$\SI{20}{\kilo\nothing} images), the former model achieves a root mean squared error (RMSE) of 1.4 pixels in detecting the four corner positions, while the latter marks 6.3 pixels.

As our final deployment platform and in-field demonstrator, we employ a commercial off-the-shelf (COTS) Crazyflie 2.1 nano-drone equipped with a Himax QVGA camera and the Greenwaves Technologies (GWT) GAP8 parallel ultra-low-power (ULP) microcontroller unit (MCU), where we run the DL front-end, while we deploy the IBVS back-end on the STM32 MCU available on the stock nano-drone.
Our closed-loop autonomous system not only reaches a real-time throughput of \SI{30}{\hertz} and \SI{27}{\hertz} for the CNN and FCNN, respectively, but compared with complex SoA systems for autonomous drone racing~\cite{de2021artificial,foehn2022alphapilot,yang2024lightweight,li2020visual}, it does not require any a priori environment map.

Finally, we field-test both pipelines, i.e., with the two alternative front-ends, stressing both endurance (up to the $\sim$\SI{4}{\minute} battery lifetime), letting the nano-drone fly in a circuit with two gates, and assessing the system capability in approaching the gate from three different orientations.
The best-performing pipeline, in both tests, is the one based on the PULP-Frontnet CNN, with a remarkable best run (out of five) of $\sim$\SI{100}{\meter} of traveled distance, never crashing, and passing through 15 gates at a peak flight speed of \SI{1.9}{\meter/\second}.
Instead, the model based on the FCNN, in its best run, covers $\sim$\SI{85}{\meter}, traversing 13 gates, with a peak flight speed of \SI{1.5}{\meter/\second}.
As the last test, we challenge our CNN-based system by deploying it in a \textit{never-seen-before} environment where it flies for more than \SI{4}{\minute}, never crashing and passing through 7 gates.


\section{Related work} \label{sec:related}

In the last decade, aerial drone races have pushed researchers to develop more agile cyber-physical systems, precise control strategies, and reactive perception pipelines up to human-level~\cite{kaufmann2023champion}.
Traditionally, this type of competition encompassed standard and micro aerial vehicles~\cite{jung2018direct, kaufmann2019beauty, foehn2022alphapilot} (from \SI{}{\kilo\gram}-scale to sub-\SI{}{\kilo\gram} robots) until recent times when also nano-sized drones, i.e., sub-\SI{50}{\gram}, entered this field~\cite{lamberti2024sim, kalenberg2024stargate, li2020autonomous}.
Static and dynamic obstacle avoidance (OA), 
and gate-based navigation (GN), i.e., passing through open frames as in Fig.~\ref{fig:demo}B, are two of the most common tasks of many drone racing competitions~\cite{lamberti2024sim, jung2018direct, kaufmann2019beauty}.

Prior knowledge of the environment map is another key aspect, while many systems assume and rely on this knowledge~\cite{jung2018direct, kaufmann2019beauty, li2020visual}, not all competitions, as well as not all real-life missions, allow this assumption.
A last fundamental characteristic of racing drones is the type of algorithm employed for their perception and control functionalities.
The two main classes are geometry-based~\cite{yang2024lightweight, li2020visual} and DL-based~\cite{lamberti2024sim, pham2022pencilnet} visual pipelines to compute the pose/position of the gates, which, to some extent, define the type of onboard sensors and computing devices needed.
Consequently, tiny and lighter DL-based workloads are more favorable for miniaturized nano-drones' ultra-constrained computational and memory budget~\cite{lamberti2024sim, kalenberg2024stargate}.
Table~\ref{tab:soa} summarizes all these aspects for many SoA autonomous systems for both micro and nano-sized classes of aerial vehicles.

Looking at the micro-sized vehicles addressing the GN task, Jung et al.~\cite{jung2018direct} presented a color-based snake gate detection algorithm~\cite{li2020autonomous} from classical computer vision.
Kaufmann et al.~\cite{kaufmann2019beauty} employed a CNN to regress the mean and the variance of the next gate's pose from RGB images at $\sim$\SI{10}{Hz}.
Recent studies~\cite{de2021artificial, foehn2022alphapilot} used the U-Net DL model~\cite{ronneberger2015u}, a FCNN similar to our baseline~\cite{crupi2024high}, which generates feature maps.
In~\cite{de2021artificial}, the authors used U-Net to segment the gates and extract the inner corner of the next gate using the snake gate algorithm, while in~\cite{foehn2022alphapilot}, the authors used U-Net in combination with part affinity fields~\cite{cao2017realtime} to detect the next gate.
Pham et al.~\cite{pham2022pencilnet} proposed a CNN to detect gates while mitigating the sim-to-real gap with an image pre-processing stage, i.e., pencil filter, which sharpens the image's edges and therefore reduces the photometric differences between synthetic images and real ones.

Moving into the nano-sized drone class size, recent works addressed the GN task~\cite{li2020visual, kalenberg2024stargate}.
In~\cite{li2020visual}, the authors employed a \SI{72}{\gram} nano-drone equipped with a powerful COTS JeVois A33 smart camera featuring a \SI{1.3}{\mega pixel} RGB image sensor, a quad-core ARM Cortex-A7, and a dual-core MALI-400 GPU, resulting in a processing power consumption of \SI{3.5}{\watt}, which is $\sim$35$\times$ higher than our power envelope.
With all this computational power, the authors could afford a novel sensor fusion method called visual model‐predictive localization, which approximates the error between the model prediction position and the visual measurements as a linear function and a RANSAC optimization stage.
Despite the complexity of this approach, this system still requires the environment's map in advance.
Instead, in~\cite{kalenberg2024stargate}, the authors presented a nano-drone capable of avoiding collision and navigating through gates, relying on two tiny CNNs running aboard fed with a low-resolution camera and a power-hungry Time-of-Flight (ToF) depth sensor, which consumes 150$\times$ the power consumption of our camera, and limits the system throughput to its maximum, i.e., \SI{15}{\hertz}.

Finally, considering the winning system of the last IMAV ``Nanocopter AI Challenge''~\cite{lamberti2024sim}, it addressed only the OA task, as the competition rules did not strictly require to pass through gates, but considered them as a bonus on the total traveled distance, which was the metric to maximize.
Therefore, the authors developed a reactive OA system based on the PULP-Dronet CNN~\cite{lamberti2022tiny}, capable of dodging dynamic obstacles and running up to \SI{30}{frame/\second} on the GWT GAP8 MCU.
By contrast, our system addresses the GN task, relying on the same hardware employed during the IMAV competition.
Starting from two different SoA DL models, i.e., the PULP-Frontnet CNN~\cite{palossi2021fully} for human pose estimation and a lightweight FCNN~\cite{crupi2024high} for drone pose estimation, we fed a classical IBVS pipeline (running aboard the nano-drone up to \SI{30}{\hertz}) to solve the GN task, relying only on monochrome low-resolution images and without any prior knowledge of the environment.
\section{System implementation} \label{sec:system}

\subsection{Hardware and software background}

\textbf{Hardware.}
We employ the COTS Crazyflie 2.1 nano-drone (Fig.~\ref{fig:demo}A), extended with the two companion boards, the AI-deck and the Flow-deck.
The stock drone features an STM32 MCU as the main flight computer, in charge of the state estimation, the low-level control with a PID cascade, and power distribution to the motors.
The AI-deck features a forward-looking Himax HM01B0 monochrome QVGA camera, an ULP GWT GAP8 System-on-Chip (SoC), enabling real-time neural network inference, and an ESP32 MCU for wireless communication.
The GAP8 and the STM32 communicate through the UART interface.
The Flow-deck enables stable flight thanks to its VL53L1x ToF laser-based height sensor and the PMW3901 optical-flow camera to track longitudinal and transversal displacements.

The GAP8 has 8+1 general-purpose RISC-V cores split into two power domains: the fabric controller (FC) and the cluster (CL). 
The FC acts as a host for the computational-intensive parallel kernels (e.g., convolutions) offloaded to the CL.
All 8+1 cores share a \SI{512}{\kilo\byte} L2 memory, while the CL has a low-latency L1 scratchpad memory of \SI{64}{\kilo\byte} which is fed by the internal direct memory access (DMA) engine.
Due to the lack of floating-point units, all computations are executed in int8 format, therefore requiring a quantization stage of the DL models.

\textbf{Software.}
Our work leverages two SoA DL models, the PULP-Frontnet~\cite{palossi2021fully} CNN and a FCNN~\cite{crupi2024high}.
The first is a feed-forward model consisting of 3 blocks featuring sequences of convolution, batch normalization, and ReLU, followed by a final fully connected layer.
This CNN performs the regression task of relative human pose estimation, given a 160$\times$96 pixel input image.
On the GAP8, this CNN processes \SI{14.1}{\mega MAC} per frame, achieving a throughput of \SI{48}{\hertz}.
The FCNN is composed of 7 convolutional blocks, each composed of a sequence of convolutional layers, batch norm, and ReLU.
Blocks 2, 3, and 4 have, in addition, a 2$\times$2 max-pooling layer that splits into half the output tensor width and height dimensions.
The task addressed by the original model is the pose estimation of a flying nano-drone; therefore, from an input image of 160$\times$160 pixels, it predicts three 20$\times$20 pixels output maps representing the depth, the position in the image space, and the target drone LED's state (on/off).
Unlike CNNs, FCNNs are not globally aware, meaning that each output pixel depends only on a subset of input pixels: the receptive field (RF).
The RF of the original FCNN is 40$\times$40 pixels.
The FCNN achieves a throughput of \SI{39}{\hertz} performing \SI{4.4}{\mega MAC} operation per frame.

\begin{figure}[!t]
    \centering
    \includegraphics[width=1.0\linewidth]{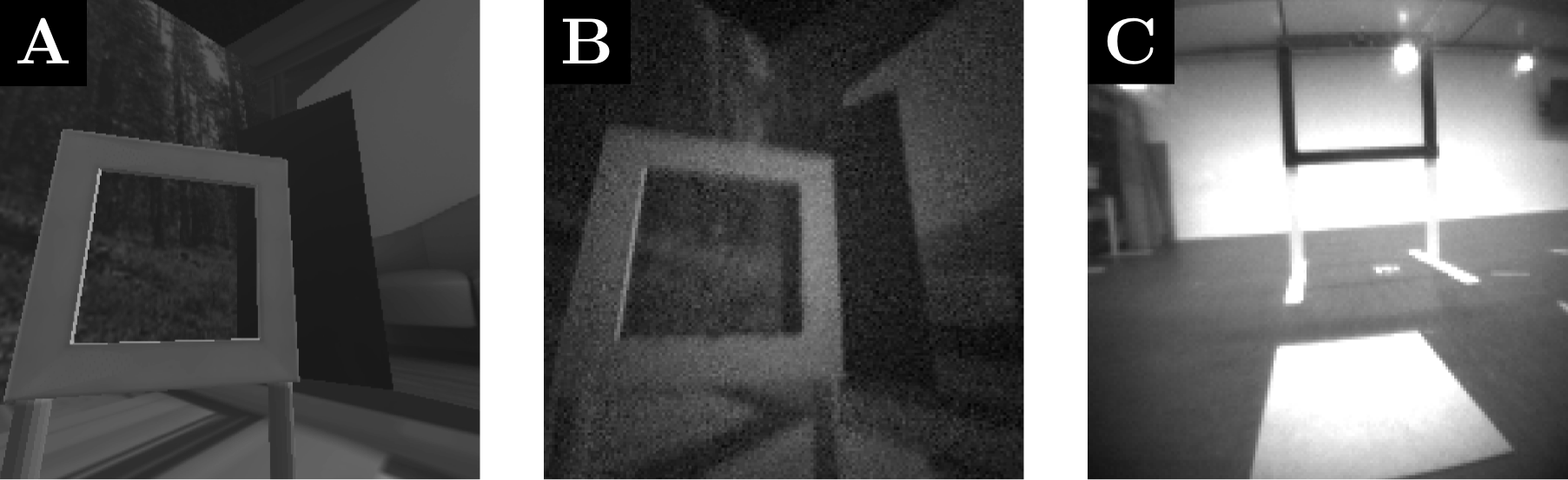}
    \caption{Examples of images collected in the simulator (A), with photometric augmentations (B), and in-field (C).}
    \label{fig:dataset}
\end{figure}

\subsection{Dataset and models training} \label{sec:SI_dataset}

\begin{figure*}
    \centering
    \includegraphics[width=1\linewidth]{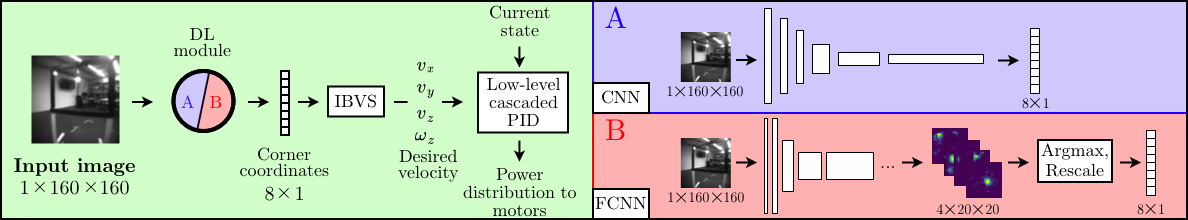}
    \caption{Our closed-loop vision-based pipeline. On the GAP8, we execute either the CNN or the FCNN; then, the 4 corner predictions are forwarded to the IBVS high-level controller, running on the STM32, which computes the desired velocities for the low-level PID cascade.}
    \label{fig:pipeline}
\end{figure*}

We build our training dataset by collecting images of the gate with various backgrounds and lighting conditions in the Webots simulator\footnote{\href{https://cyberbotics.com}{\url{https://cyberbotics.com}}} by randomly spawning the drone (Fig.~\ref{fig:dataset}A).
To increase the dataset variability, we use geometric augmentations, like random flip, random rotation, random erasure, and scale-increasing, along with photometric augmentations, such as motion blur, gaussian blur, vignetting effect, and random noise and a custom procedure to mimic the actual onboard camera, reducing the sim-to-real gap.
The final synthetic dataset used for training is thus composed of \SI{750}{\kilo\nothing} augmented images (Fig.~\ref{fig:dataset}B), of which  \SI{600}{\kilo\nothing} are used to train the two neural networks, while the remaining \SI{150}{\kilo\nothing} compose the validation set. 

A smaller dataset of real-world images (Fig.~\ref{fig:dataset}C) is collected in our laboratory equipped with an OptiTrack motion capture (mocap) system.
Our setup consists of one gate faced by the nano-drone sweeping the room following a 1$\times$\SI{1}{\meter} grid path at three different heights: 0.5, 1.0, and \SI{1.5}{\meter}.
During the data collection, the drone starts from 3 relative orientations w.r.t. the gate, i.e., $-45\degree, 0\degree, \text{and} +45\degree$.
We collect images from various positions, capturing all four walls of the laboratory, with a number of samples ranging from 23\% to 27\%.
Moreover, we include images featuring partial views of the gate.
In the majority of the samples (83.75\%), the four corners of the gate are always visible, while the remaining samples have three (2.6\%), two (11.88\%), one (1.72\%), or zero (0.05\%) visible corners.
The final dataset has $\sim$\SI{100}{\kilo\nothing} images, divided into the training, validation, and testing set of $\sim$\SI{70}{\kilo\nothing}, $\sim$\SI{10}{\kilo\nothing}, and $\sim$\SI{20}{\kilo\nothing} samples, respectively.
In both simulation and real-world datasets, the ground truths (GTs) are computed by projecting the 3D positions of the corners into the image plane using the pinhole camera model.
While the CNN can directly use these values as labels, for the FCNN, these GTs are further transformed into feature maps, one for each corner~\cite{crupi2024high}.

Considering the perception module, we modified the PULP-Frontnet's architecture to process $160\times160$ images and to produce eight scalars corresponding to the image coordinates of the four corners (Fig~\ref{fig:pipeline}), while in the FCNN, we also increased its RF.
By substituting the convolutional layer of blocks 4 and 5 from 3$\times$3 to a 5$\times$5 and a 7$\times$7 convolutional kernels, respectively, we increased the FCNN's RF to 96$\times$\SI{96}{px}.
This modification allows the FCNN to include additional image features, such as the gate's legs, which results in improved detection performances, as shown in Sec.~\ref{sec:results_perception}.
Our FCNN outputs 4 feature maps, one for each corner, which are further processed with \texttt{argmax} operator to extract the coordinates of the corners and rescaled to $[0,159]$.

Both the CNN and the FCNN are trained on synthetic data for 100 epochs with early stopping criteria of 10 epochs, a learning rate of 0.001, and the Adam optimizer. 
They are both fine-tuned on real-world images for 20 epochs with an early stopping of 5 epochs.
The CNN is trained using the L1-loss; the FCNN uses the mean squared error as in~\cite{foehn2022alphapilot}.

\subsection{Closed-loop vision-based pipeline} \label{sec:SI_closedloop}

The control module is fed with the eight corner coordinates coming from the DL one (either CNN or FCNN).
It comprises an IBVS controller that computes the drone velocity commands, which the low-level controller uses to set the servo motors moving the drone's propellers.
Regarding the IBVS controller, we adapt its simple formulation~\cite{chaumette2006vs} that computes the drone's velocity as $\bm{v} = \lambda \bm{L}^+(\bm{p}^*-\bm{p})$ using as feedback the image information expressed in form of visual features $\bm{p}$.
In this reactive law, $\lambda$ is a positive gain governing the converge rate of the control action.
Instead, $\bm{L}^+$ is the pseudo-inverse of the Jacobian matrix that relates the drone's velocity to the features' time derivative. 
The Jacobian normally depends on the features' depth that, to alleviate the computational load from possible estimation procedures, we set constant to \SI{0.5}{\meter}.
Furthermore, to cope with the underactuated nature of drones~\cite{Corke:book:2023}, the Jacobian does not include the columns related to the roll and the pitch rate.
This means that the velocity command is composed of the 3D linear and vertical angular velocity (as in Fig.~\ref{fig:pipeline}). 

In our scheme, the visual features vector $\bm{p}$ is filled with the coordinates of the four corners measured online by the DL module. 
The desired features in $\bm{p}^\ast$ are set in advance with the value that the features would have when the drone is in front of the gate at a distance of \SI{0.5}{\meter}. 
Thus, as the IBVS aims to zero the difference $\bm{p}^\ast-\bm{p}$, the control action drives the drone in front of the gate using its current measure, accounting for
possible variations of the gate pose.

\subsection{Deployment} \label{sec:SI_deployment}

To enable efficient integer arithmetic on the GAP8 SoC, we perform a post-training quantization-aware fine-tuning from float32 to int8 data type using the QuantLib\footnote{\href{https://github.com/pulp-platform/quantlab}{\url{https://github.com/pulp-platform/quantlab}}} open-source tool.
Then, we used DORY~\cite{burrello2021dory} to automatically generate the C code running on the GAP8. 
DORY generates code with all the DMA transfers between all the memory levels of the GAP8 by exploiting the PULP-NN~\cite{garofalo2020pulp} library. 
In the closed-loop pipeline, the perception module runs on the GAP8, while the IBVS and the low-level control are executed on the STM32.
The perception module processes the images and extracts the corners' coordinates, which are sent through the UART interface to the STM32, where the IBVS computes the velocity commands, feeding the cascaded PID controller.
\section{Results} \label{sec:results}

\subsection{Offline analysis of the perception models} \label{sec:results_perception}

Table~\ref{tab:offline_test} compares our two DL models against the original FCNN~\cite{crupi2024high} and a dummy predictor, which always predicts the mean values of the testing set.
We report the number of parameters for each model, the number of MAC operations per frame, the memory footprint, and the RMSE of the gate detection for both float32 and int8 models.
Our results show how quantization does not affect the performance of any model.
Considering the int8 version, we can see how the larger CNN is the best-performing model with an RMSE of \SI{1.45}{px}.
Instead, the comparison between the original FCNN and our version shows the benefit of the increased RF, resulting in a reduction of the RMSE of $\sim$40\%.
Finally, considering the closed-loop performance of the two models running on the GAP8 (FC@\SI{240}{\mega\hertz}, CL@\SI{175}{\mega\hertz}), the CNN marks a higher throughput, i.e., 30 vs. \SI{27}{frame/\second}.

\begin{table}[!t]
\centering
\caption{Offline performance of the perception models}
\label{tab:offline_test}
\begin{tabular}{cccccc}
\toprule
\multirow{2}{*}{Model} & Params. & \multirow{2}{*}{MMAC} & {Memory} & \multicolumn{2}{c}{Mean RMSE {[}px{]}} \\ 
 & [k] &  & [MB] & float32   & int8 \\ 
\midrule
CNN  & 322 & 23.87 & 5.14 & 1.40 & 1.45 \\
FCNN & 38 & 23.3 & 3.99   & 6.28 & 6.31 \\
Crupi et al.~\cite{crupi2024high} & 6.5 & 7.90 & 3.86 & 10.40 & - \\
Dummy & - & - & - & 25.21 & - \\
\bottomrule
\end{tabular}
\end{table}

\subsection{In-field closed-loop experiments}

\begin{figure*}
    \centering
    \vspace{2mm}
    \includegraphics[trim={0cm, 0, 0 ,0},clip=true,height=0.210\linewidth]{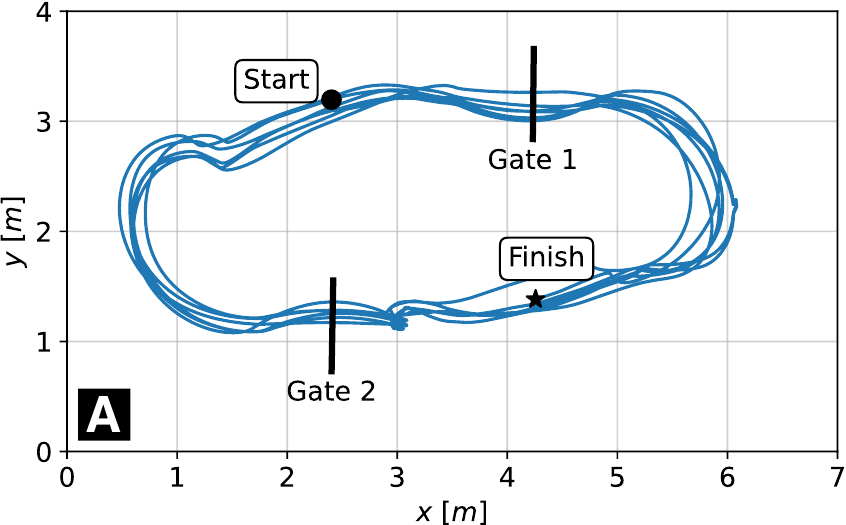}
    \hfill\includegraphics[trim={1cm, 0, 0 ,0},clip=true,height=0.210\linewidth]{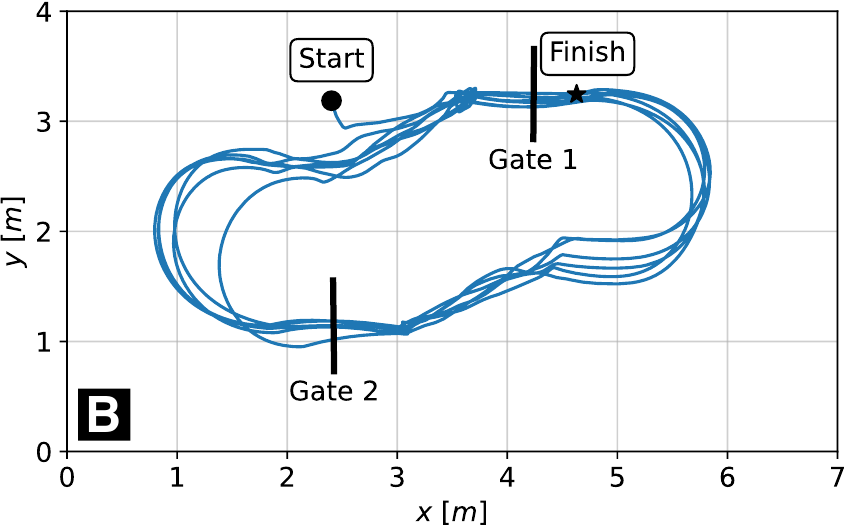}
    \hfill\includegraphics[trim={1cm, 0, 0 ,0},clip=true,height=0.210\linewidth]{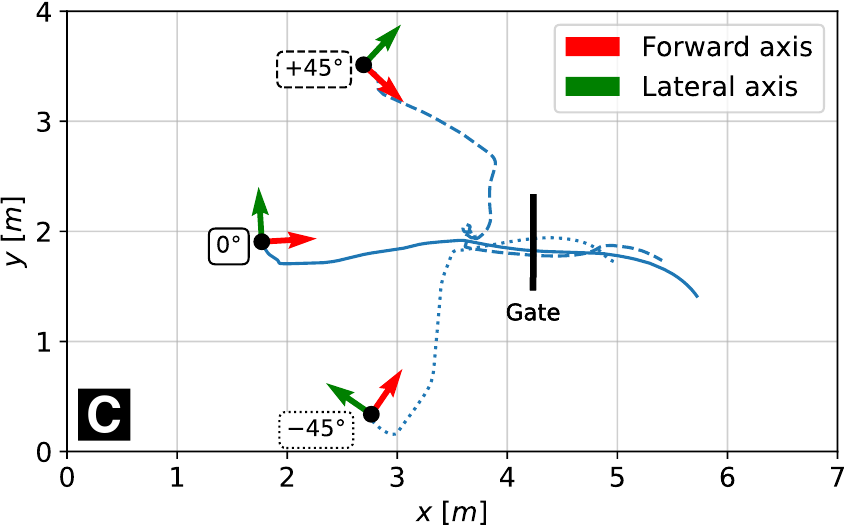}
    \caption{Sample trajectories of the first in-field experiment, either employing the CNN (A) or the FCNN model (B). In (C), we show an example of the second in-field experiment where the nano-drone faces the gate from three different initial positions.}
    \label{fig:experiments_traj}
\end{figure*}

In our first in-field experiment, we task the nano-drone to continuously navigate through two static gates.
We record the nano-drone motion with a mocap system (not used for perception or control).
We implement a state machine consisting of the following states: the drone \textit{take-off}, once it reaches \SI{1}{\meter} height, it passes to the \textit{gate-based navigation} state where it approaches the gate until the IBVS error reaches \SI{8}{px}, then it enters in an open-loop procedure \textit{move forward and turn}.
In this last procedure, the drone performs a pre-defined \textit{forward movement} of \SI{1.0}{\meter} and \SI{180}{\degree} turn, then it switches back to the closed-loop \textit{gate-based navigation} task.
In this test, the IBVS's $\lambda$ gain is set to 0.5.
Table~\ref{tab:endurance} reports the performance of the two models out of 5 runs. 
We consider a run successful if the drone continuously passes through gates, without collisions, for the entire duration of its battery (i.e., $\sim$\SI{4}{\minute}).
The CNN-based pipeline performs better than the FCNN-based one, having a higher mission success rate, i.e., 4 runs out of 5, traversing more gates, and traveling for a longer distance \SI{87}{\meter} vs. \SI{54}{\meter}, on average. 
Figure~\ref{fig:experiments_traj}A shows a sample run using the CNN, while Fig.~\ref{fig:experiments_traj}B shows an example with the FCNN perception module.

\begin{table}[t]
\centering
\caption{First sets of experiments: comparison of the pipeline using the two perception models during long-lasting experiments}
\label{tab:endurance}
\begin{tabular}{cccccc}
\toprule
\multirow{2}{*}{Model} & {Mission} & \multicolumn{2}{c}{Gates passed} & \multicolumn{2}{c}{Distance {[}m{]}} \\
& success rate & Average & Best & Average & Best \\
\midrule
CNN  & 4/5 & 13 & 15 & 87 & 97 \\
FCNN & 2/5 & 8  & 13 & 54 & 85 \\ 
\bottomrule
\end{tabular}
\end{table}

The second experiment stresses the robustness of our systems against the different angles of approaching the gate.
The drone faces the gate with a variable relative orientation among \SI{-45}{\degree}, \SI{0}{\degree}, and \SI{+45}{\degree}, which also implies a different distance from the gate.
For each configuration, we repeat the test five times and define the gate traversal success rate as the ratio between the number of times the drone successfully traverses the gate and the total number of flights.
Fig.~\ref{fig:experiments_traj}C shows the three configurations and actual trajectories recorded during the tests with the CNN-based pipeline.
Tab.~\ref{tab:orientation} summarizes the results of this second experiment, showing how, once again, the CNN-based pipeline has an almost-ideal behavior, successfully navigating through the gate 14 times out of 15 (5 runs for each orientation).
Furthermore, the pipeline using the CNN model requires less time to accomplish the task than the counterpart using the FCNN.
The FCNN-based pipeline consistently fails in passing through the gate when approaching it with a relative orientation of \SI{-45}{\degree}.
In this case, the more challenging background leads to noisy predictions of the FCNN, resulting in aggressive velocity commands from the IBVS, thus losing the gate from the field of view (FoV). 

\begin{table}[!t]
\centering
\caption{Second set of experiments: comparison of the pipeline using the two perception models varying the initial conditions.}
\label{tab:orientation}
\begin{tabular}{ccccc}
\toprule
\multirow{2}{*}{Model} & {Initial} & {Gate traversal} & \multicolumn{2}{c}{Time {[}s{]}} \\
& {Orientation {[}deg{]}} & Success rate & Average  & Best  \\ \midrule
\multirow{3}{*}{CNN}   & +45  & 5/5  & 29 & 16 \\
& 0  & 5/5 & 20  & 13  \\
& -45 & 4/5 & 16 & 15 \\ \midrule
\multirow{3}{*}{FCNN}  & +45 & 3/5 & 31  & 25 \\
& 0 & 5/5 & 26   & 16  \\
& -45  & 0/5  & - & -  \\ \bottomrule
\end{tabular}
\end{table}

\begin{figure}
\centering
\includegraphics[width=\linewidth]{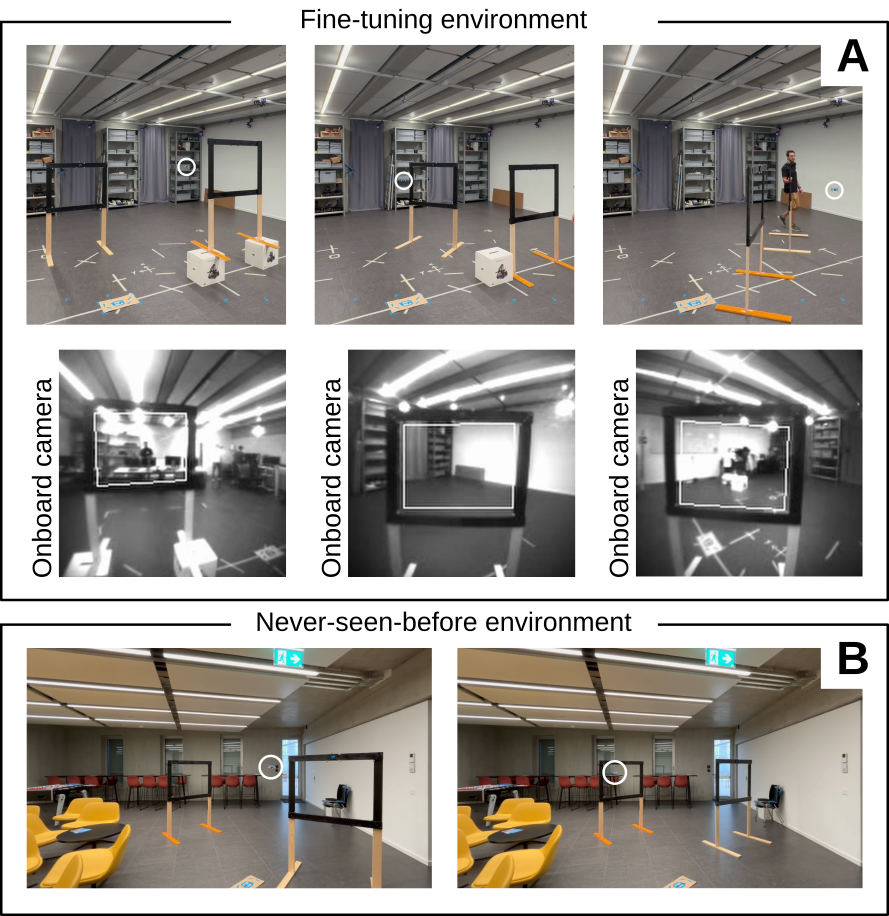}
\caption{Footage of the in-field experiments, either in our laboratory (A), i.e., the fine-tuning environment, or in a never-seen-before room (B).}
\label{fig:snapshots}
\end{figure}

Then, we assess the drone's ability to navigate in a challenging environment where the gates are continuously moved (including vertically), even when the drone is approaching them. 
We employ the best-performing CNN model, and we let the drone fly for \SI{5}{\minute} (over the average battery duration of \SI{4}{\minute}) in our laboratory with two gates (Fig.~\ref{fig:snapshots}A).
We exploit the same state machine described above, ensuring gates are, at least partially, within the drone's FoV.
The nano-drone achieves a total travel distance of \SI{108}{\meter}, passing 13 times through the gates and never crashing.
Finally, to stress the generalization capability of our DL model, we repeat the same test in a \textit{never-seen-before} environment (Fig.~\ref{fig:snapshots}B), where everything, except for the gates, differs from both training and fine-tuning data.
Despite CNN predictions being less accurate than in our laboratory, our drone navigates through the gates 7 times, flying for more than \SI{4}{\minute} before losing track of the gates.
We provide a supplementary video showing both experiments\footnote{\href{https://youtu.be/aY0dIhoKQXg}{\url{https://youtu.be/aY0dIhoKQXg}}}.
 
\subsection{Discussion and future work}

Despite the remarkable results achieved by our systems in the GN task, our system assumes that a gate is always, at least partially, visible from the drone, which we enforce in our experiments by a controlled position of the two gates.
A simple solution to this limitation can be achieved by extending the perception module with a binary classifier that predicts the presence of a gate.
If no gate is detected, the nano-drone can spin in place until a new gate is found.

Additionally, our system does not have an obstacle avoidance functionality, which is paramount for any premiere drone race.
By leveraging the CNN for reactive obstacle avoidance used in~\cite{lamberti2024sim} (winner of the IMAV'22 nano-drone race) and deployed on the same GAP8 SoC used in our work, we could envision a perception pipeline coupling it, in time-interleaving, with our DL module.
However, this solution would half the overall throughput, i.e., \SI{15}{\hertz}, as both our model and the one for obstacle avoidance run at $\sim$\SI{30}{frame/\second}, marginally limiting the robot's performances.
An appealing alternative is represented by the Tiny-PULP-Dronet CNN~\cite{lamberti2024distilling} for obstacle avoidance, which is extremely light (only $\sim$\SI{1}{\mega MAC/frame}), and peaks at~\SI{139}{frame/\second} on our same hardware, resulting more than 5$\times$ faster compared to~\cite{lamberti2024sim}.
Combining this tiny CNN with ours would marginally limit the final throughput, i.e., $\sim$\SI{25}{frame/\second}.
However, the tiny model has never been proven in a nano-drones race, leaving open the question of its effectiveness in preventing collisions in such a challenging scenario.

Lastly, the most interesting evolution in this context would be to design a novel DL brain addressing both tasks with a shared CNN's back-bone and two separate heads, where the first would predict one~\cite{lamberti2024distilling} or multiple~\cite{lamberti2024sim} probabilities of collision, and the latter would predict the gate position.
This research direction is now possible thanks to our system, which is, to the best of our knowledge, the first only-vision (monocular camera) DL-based framework for reactive GN aboard nano-drones.
\section{Conclusion} \label{sec:conclusion}

We present two pipelines for monocular gate-based navigation, a typical task of drone-race competitions.
Our pipelines are tailored to run in real-time (up to \SI{30}{frame/\second}) on the limited resources aboard the Crazyflie 2.1 nano-drone.
Each pipeline comprises one of the two SoA DL-based front-ends we adapted for our task (model modifications, training, fine-tuning, and quantization), in charge of detecting the four corners of the gate, and an IBVS back-end feeding the low-level controller with linear and angular velocities.
Our in-field results, show \textit{i}) a peak performance of $\sim$\SI{100}{\meter} travelled distance, in \SI{4}{\minute}, at a peak speed of \SI{1.9}{\meter/\second}, passing trough 15 gates, and \textit{ii}) the capability of our system to successfully fly for more than \SI{4}{\minute} in a never-seen-before challenging environment.
Compared to other SoA solutions, our system is the first, to the best of our knowledge, to accomplish the gate-based navigation task on a nano-drone, requiring neither a depth sensor nor a priori knowledge of the environment.


\addtolength{\textheight}{-7.2cm}  


\bibliographystyle{IEEEtran}
\bibliography{bibliography.bib}

\end{document}